\documentclass{article}

\usepackage{microtype}
\usepackage{graphicx}
\usepackage{subfigure}
\usepackage{booktabs} 
\usepackage{amsmath}
\usepackage{caption}
\usepackage{multicol} 
\usepackage{placeins} 
\usepackage{stfloats} 

\usepackage{subcaption}

\usepackage{hyperref}

\usepackage{xurl}

\usepackage[accepted]{mlsys2025}
\makeatletter
\renewcommand{\mlsys@appearing}{Preprint.}
\makeatother

\mlsystitlerunning{Tiered Cache Blocking and In-Kernel Task Dispatch for Long-Context Prefill Attention on CPU}

\begin{document}

\twocolumn[
\mlsystitle{Tiered Cache Blocking and In-Kernel Task Dispatch for Long-Context Prefill Attention on CPU}



\mlsyssetsymbol{second}{$\dagger$}

\begin{mlsysauthorlist}
\mlsysauthor{Kwun Fung Lau}{intel}
\mlsysauthor{Frost Mitchell}{second,intel}
\mlsysauthor{Kanya Mo}{second,intel}
\mlsysauthor{Emani Ashok}{intel}

\end{mlsysauthorlist}

\mlsysaffiliation{intel}{Intel Corp., USA}
\mlsyscorrespondingauthor{Kwun Fung Lau}{kwun.fung.lau@intel.com}

\mlsyskeywords{Machine Learning, Attention, Prefill, CPU, Agentic AI, Inference}

\vskip 0.3in

\begin{abstract}
Agentic AI drives significant productivity gains across diverse domains, but concurrent long-context prefills from multiple agents can saturate GPU compute resources. Although CPUs offer a viable alternative, existing CPU attention kernels fall short in adapting to multi-tier cache hierarchies and lack effective thread scheduling. To address this, we introduce TACPU, a high-performance attention kernel featuring tiered cache blocking and an in-kernel task dispatching mechanism designed for causal full-sequence prefilling. TACPU achieves up to a 2.01× speedup over PyTorch SDPA, FlexAttention, and SGLang attention kernels. Integrated into SGLang, it reduces Time to First Token (TTFT) of Llama-3.1-8B by up to 1.74× while preserving model accuracy.

\end{abstract}
]



\printAffiliationsAndNotice{\textsuperscript{$\dagger$}Equal second-author contribution. }
\section{Introduction}

Several recent efforts have explored CPU-based acceleration for large language models
(LLMs). \citet{gupta2026flashattention} adapt FlashAttention to scalable vector
architectures, so that the kernel scales from short to long vectors on RISC-V CPU. ~\citet{na2025flexinfer} exploit CPU memory and computation alongside GPU resources, while ~\citet{shen2025efficient} improve CPU LLM inference through INT4 weight-only quantization and an optimized runtime.
At the framework level SGLang and vLLM have introduced CPU-specific kernels ~\citep{jaylsde_2026_avx512_sglang} and scheduler
optimizations~\citep{li_2025_cpu_vllm}. In this work, we
specifically target causal attention mechanisms to accelerate the LLM serving pipelines,
which will benefit agentic AI workloads on CPU architectures.

Agentic AI workloads require large language models (LLMs) to process extremely long contexts, which include intermediate reasoning steps~\citep{wei2022chain} and tool call outputs~\citep{schick2023toolformer}. Building upon recent advancements in GPU attention kernels~\citep{dao2022flashattention, dao2024flashattention2, shah2024flashattention3, zadouri2026flashattention4}, we propose an optimized CPU attention kernel tailored specifically for the prefill phase to accelerate long-context processing in LLM serving frameworks. We introduce two core mechanisms: Two-Level Cache-Aware Scheduling and L3-Chunk-Scoped Cost-Prioritized Scheduling. The former minimizes redundant DRAM reads by partitioning sequences according to cache hierarchy budgets, while the latter addresses core workload imbalance through dynamic cost-based task dispatch from a shared task queue. Section 3 details the algorithmic design of both mechanisms, and Section 4 presents their evaluations.

To be specific, our contributions are:
\begin{itemize}
\item We propose a cache-aware scheduling strategy to minimize DRAM traffic, improve cache locality, and reduce latency for long sequences. Additionally, we introduce an analytical model that derives appropriate hyperparameters directly from tensor memory footprints and CPU cache capacities, eliminating the need for empirical per-CPU tuning. 
\item We extend longest-processing-time-first (LPT) scheduling into the runtime kernel to dynamically balance workloads across both homogeneous and heterogeneous CPU architectures. By introducing an exact-cost analytical model and implementing a shared work queue in-kernel, our approach bypasses rigid manual thread pinning and delivers seamless portability across serving frameworks. 
\item We identify a strong interaction effect between our two proposed optimizations. Microbenchmarks and ablation studies show that cache-aware scheduling alone is counterproductive on heterogeneous cores and provides only marginal benefits on Xeon CPUs. In contrast, integrating both techniques yields significant performance improvements. This provides a design insight for high-performance CPU kernel development.

\end{itemize}

\newpage
\section{Related Work}
As the attention mechanism constitutes the primary computational backbone of Large
Language Models (LLMs), extensive research has focused on its optimization to fully
utilize GPU capabilities. A typical example is FlashAttention (FA). Beginning with
FA1~\citep{dao2022flashattention} and FA2~\citep{dao2024flashattention2}, the primary
goal was to minimize redundant data transfers between high-bandwidth memory (HBM) and
on-chip SRAM. With the advent of dedicated Tensor Cores, FA3~\citep{shah2024flashattention3}
extended this concept by overlapping General Matrix Multiplication (GEMM) with Softmax
operations through asynchronous execution and warp specialization. Most recently,
FA4~\citep{zadouri2026flashattention4} co-designs the algorithm with the kernel pipeline
to address the asymmetric hardware scaling of Blackwell-class systems. These
methods yield exact outputs while significantly reducing latency.


\subsection{Long Context Scenarios}
Specialized attention mechanisms have also been developed to tackle long-context
scenarios efficiently. Local (sliding-window) attention reduces attention complexity
to process long documents~\citep{beltagy2020longformer}; a related work also
shows that biasing attention toward local image regions can also mitigate hallucination in
vision-language models~\citep{an2025mitigating}.
Infini-Attention~\citep{munkhdalai2024leave} extends this concept by incorporating a
compressed memory module to support streaming over unbounded context lengths with memory
constraints, while \citet{xiao2024efficient} retain only a fixed window together with
the initial attention sinks. Other approaches focus on distributed compute efficiency:
RingAttention with Blockwise Transformers~\citep{liu2024ringattention} overlaps
peer-to-peer (P2P) Key/Value communication with computation in a logical ring network.
To address the inherent workload imbalance in causal attention, Striped
Attention~\citep{brandon2023striped} redistribute blocks across accelerators to
increase hardware utilization.

\subsection{Serving Frameworks}
To satisfy the demands of high-throughput LLM serving systems, several framework-specific attention primitives have emerged. vLLM introduced
PagedAttention~\citep{kwon2023efficient}, which applies virtual memory paging principles to dynamically manage KV cache allocations. Conversely, SGLang adapts a RadixTree structure to efficiently manage and reuse shared prefix KV caches~\citep{zheng2024sglang}. FlashInfer~\citep{ye2023flashinfer} unifies these different KV cache layouts under a single hardware-optimized kernel abstraction.

\subsection{Offloading Workloads to CPU}
Recent work actively leverages CPU resources to mitigate the massive computational demands of LLM serving. FlexGen~\cite{flexgen} offloads attention-score computation to the CPU when the KV cache resides in CPU memory, while FastDecode~\cite{fastdecode} extends this approach by executing KV-cache-bound attention across aggregated cross-node CPU compute and memory bandwidth. For online serving, NEO~\cite{neo} overlaps CPU attention with GPU computation, transmitting only compact activations rather than large KV caches. Other approaches partition a single attention operation across devices: HGCA~\cite{hgca} splits dense and sparse attention between the CPU and GPU before merging them, while HybridGen~\cite{hybridgen} parallelizes attention logits using a feedback-driven dynamic scheduler.

\subsection{Variants of Attention Mechanisms on CPUs}
Several efforts focus on optimizing CPU-native attention performance through either structural redesign or approximate computation. For architectural modifications, NOSA~\cite{nosa} applies trainable sparse attention to bound KV cache transmission. For approximation methods, IceFormer~\cite{iceformer} approximates attention using top-k candidate keys selected by nearest-neighbor search, while MagicPIG~\cite{magicpig} samples keys and reweights them according to their sampling probabilities. Alternatively, NoMAD-Attention~\cite{nomad} approximates matrix multiplication by turning multiply-add operations into in-register SIMD lookups, IntAttention~\cite{intattention} numerically approximates softmax by clipping and using lookup tables with integer normalization. SparAMX~\cite{sparamx} prunes low-magnitude KV values to accelerate GEMMs using hardware-accelerated Intel AMX kernels.

\subsection{Our Work}
While these pioneering approaches highlight promising research directions, many require model retraining, model architecture redesigns, or active framework adaptations. These introduce difficulties to adopt in existing agentic AI workloads. In contrast, this work aims to deliver immediate and practical performance gains for existing LLM serving systems. Specifically, we introduce a self-contained attention kernel that preserves output quality while seamlessly integrating into mainstream frameworks such as PyTorch and SGLang. By maintaining full compatibility with pre-trained LLMs, our design substantially improves end-to-end prefill throughput in modern production environments.

\section{Background and Design}
\label{Design}
\subsection{FlashAttention on GPU}
FlashAttention~\citep{dao2022flashattention} avoids instantiating the $N \times N$ attention matrix by tiling $Q$, $K$, and $V$ tensors into fast on-chip SRAM. To achieve this, it proposes the online softmax equation: 
\begin{equation}
\begin{aligned}
m' & \gets \max\big(m,\, \mathrm{rowmax}(S_j)\big),\\
l  & \gets e^{m-m'}\,l + \mathrm{rowsum}\big(e^{S_j - m'}\big),\\
O  & \gets e^{m-m'}\,O + e^{S_j - m'}\,V_j
\end{aligned}
\label{eq:online-softmax}
\end{equation}

where $S_j = \alpha Q_b K_j^\top$, $\alpha$ is the scaling factor, $Q_b$ is the query block, $m$ is the running row maximum, $l$ is the running normalization factor, and $O$ is the unnormalized output accumulator. 

FlashAttention-2~\citep{dao2024flashattention2} further improves GPU throughput by pinning $O$, $m$, $l$ in registers while FlashAttention-3~\citep{shah2024flashattention3} overlaps softmax operations with asynchronous GEMMs and leverages persistent kernels with load-balanced work scheduling. FlashAttention-4~\citep{zadouri2026flashattention4} targets Blackwell's compute-memory asymmetric scaling via emulated exponentials and longest-processing-time-first (LPT) ~\citep{graham1969lpt} causal query scheduling.

While these GPU-centric techniques yield substantial speedups, fundamental
architectural differences prevent them from mapping efficiently to modern CPUs. To address this gap, we propose two key mechanisms tailored specifically for CPU execution.


\subsection{Two-Level Cache-Aware Scheduling}
Unlike GPU shared memory, CPU L2 and L3 caches are filled by hardware according to the order of memory accesses. Data access must be carefully scheduled to preserve locality. Dedicated loop structures are hence introduced to exploit the cache hierarchy of the CPU, without changing the arithmetic of the kernel. At the entry point of the kernel, we compute the working sets  analytically from the CPU cache capacities. This proposed scheduling can be divided into 2 main parts: L3 cache and L2 cache. \newline \newline
\textbf{L3 Cache: } For short and medium length sequences, the shared cache (L3) is not a bottleneck as the key and value tensors fit entirely in L3. As the sequence grows, this stops holding. Beyond a certain sequence length, K and V blocks are evicted before the next query block reuses them, forcing repeated reads from DRAM. This results in redundant DRAM-to-L3 traffic, which is why the L3 capacity has to enter the algorithm design rather than being left to the hardware. 

\begin{figure*}[t!]
    \centering
    \includegraphics[scale=0.4]{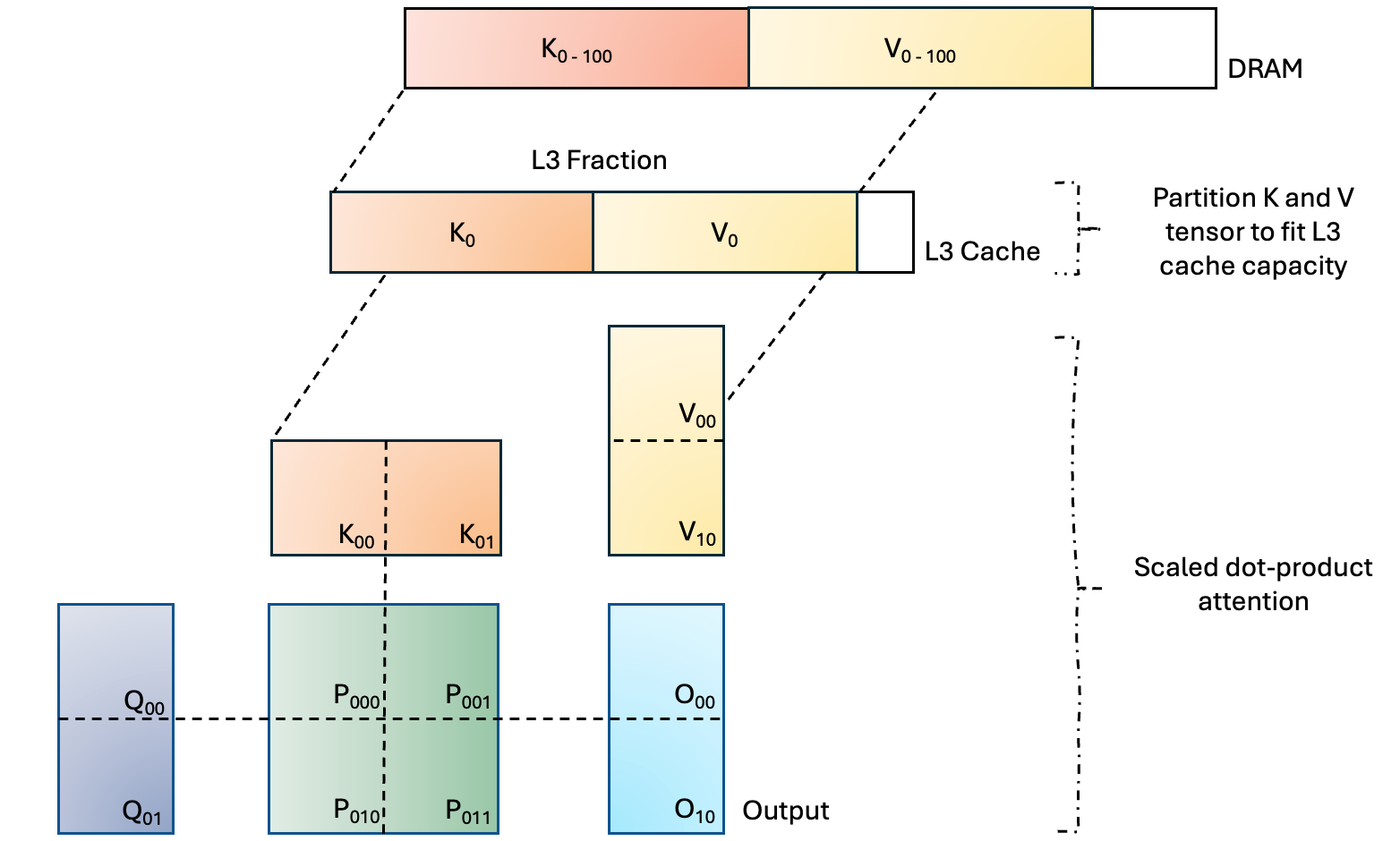}
    \caption{Illustration of K and V tensor packing from DRAM into the L3 cache}
    \label{fig:l3_cache_fig}
\end{figure*}

We restructure the loop nest so that the K and V tensors brought into L3 are fully reused before they can be evicted. We group Q heads whose K and V tensors fit within a fixed fraction of L3. An outer loop is added in order to walk the groups one at a time. If a single head's K and V already exceed the budget, we will further split the sequence as well. We parallelize the threads on the same group. Since every core on a socket shares one L3 cache, a block of K or V tensors that any thread pulls in is immediately available to all the other cores. As a result, unnecessary L3 cache eviction can be avoided, and the performance of the algorithm can be improved. Figure \ref{fig:l3_cache_fig} shows the packing of K and V tensors.

\textbf{L2 Cache:} Once the L3-resident chunk has eliminated DRAM refetch, we move one level down. Under the natural ordering, where each $q$-block sweeps the entire chunk, every KV step ($j$) is re-fetched from L3 once per $q$-block. This increases the latency of the kernel.

In order to address this, we invert the two innermost loops: the loop over KV steps becomes the outer loop, while the loop over a group of $G$ consecutive $q$-blocks becomes the inner loop. So a packed KV step ($j$) is loaded into L2 once and reused by $G$ $q$-blocks before it is evicted.

Rather than manually tuning $G$ for different CPUs, we conservatively size it from the L2
capacity. We split the group's resident set into two terms: shared ($U$) and per-$q$-block ($Q_{\text{b}}$).

\begin{equation}
\label{eq:group-formula}
\begin{aligned}
G = \frac{f_{\text{L2}} L_2 - U}{Q_{\text{b}}}
\end{aligned}
\end{equation}

where $f_{\text{L2}}$ is the L2 fraction, $L_2$ means the byte size of the L2 cache, $U$ represents the shared term, and
$Q_{\text{b}}$ denotes the per-$q$-block term.

\begin{equation}
\label{eq:l2-per-qblock}
\begin{aligned}
  Q_{\text{b}} = q_{\text{r}}(Q_{\text{d}} + V_{\text{d}} + m + l)
\end{aligned}
\end{equation}

in which $q_{\text{r}}$, $Q_{\text{d}}$ , $V_{\text{d}}$, $m$ , and $l$ denote rows of the $q$-block (i.e. a step of Q tensor), head dimension of the Q tensor, head dimension of the V tensor, row max and row sum respectively. Physically, this means the rows of each $q$-block plus the output, row max and sum should survive every KV step.

For the shared term: 
\begin{equation}
\label{eq:shared}
\begin{aligned}
  U = k_{\text{c}}(K_\text{d} + V_\text{d}) + S_\text{b}
\end{aligned}
\end{equation}

where $k_{\text{c}}$, $K_\text{d}$, $V_\text{d}$, and $S_\text{b}$ refer to columns of the k-block, head dimension of the K,
head dimension of the V, and the score and probability tiles ($S$ and $P$) for one $q$-block.

\begin{figure}[b!]
    \centering
    \includegraphics[width=\columnwidth]{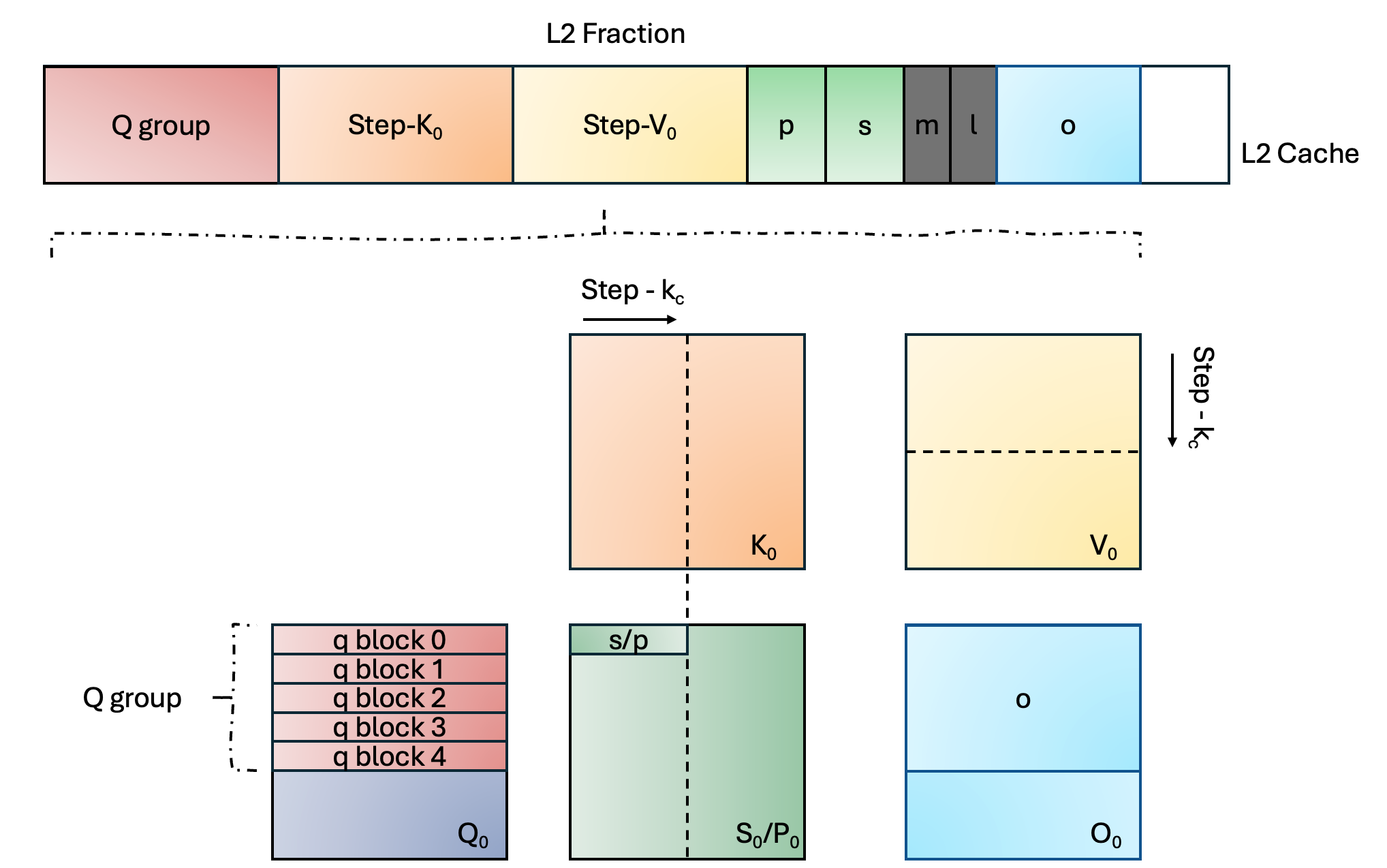}
    \caption{Data Retention in the L2 Cache}
    \label{fig:l2_cache_fig}
\end{figure}

In short, the shared term ($U$) represents one step of K and V sizes, while $Q_\text{b}$ refers to the sizes required per $q$-block.
Figure \ref{fig:l2_cache_fig} visualizes the concept of it. All terms are in bytes: each element count is multiplied by its element size (2 bytes for bf16 Q, K, V and $P$; 4
bytes for fp32 output, $m$, $l$ and $S$).

\subsection{L3-Chunk-Scoped Cost-Prioritized Scheduling}
\textbf{The load-imbalance problem:} The two-level cache-aware scheduling scheme described in the previous section establishes a specific loop hierarchy, where the KV step serves as the outer loop and the query block ($q$-block) as the inner loop. Consequently, the query group ($q$-blocks) forms the natural unit of parallelism. 

\begin{figure}[b!]
    \centering
    \includegraphics[scale=0.45]{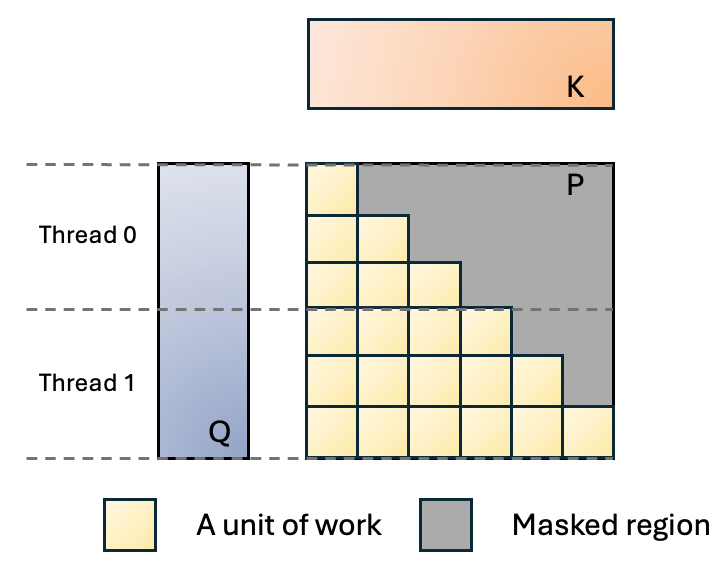}
    \caption{Workload imbalance across threads}
    \label{fig:work_imbalance_fig}
\end{figure}

In self-attention, contiguous partitioning works effectively as all threads execute identical workloads. This does not hold in causal attention. Since causal attention applies a lower-triangular mask, threads handling $q$-groups from earlier sequence positions skip fully masked regions and finish quickly. One example is Thread 0, which is shown in Figure \ref{fig:work_imbalance_fig}.  In contrast, threads assigned later $q$-groups, like the Thread 1, bear more units of work. Due to the barrier synchronization among the threads, faster threads have to wait for the slowest thread. This imbalance introduces severe execution stalls.

\textbf{FlashAttention-4 LPT Scheduling and Its CPU Limits:} FlashAttention-4~\citep{zadouri2026flashattention4} addresses work imbalance via LPT scheduling ~\citep{graham1969lpt}, assuming identical processors and monotonic tile execution costs in causal attention. Although effective for GPUs, these assumptions do not hold on modern CPUs with heterogeneous execution units and complex cache hierarchies. We adapt and build upon the LPT heuristic to address imbalanced workloads on modern CPUs.

\textbf{Proposed solution:} To extend LPT scheduling ~\citep{graham1969lpt} and integrate our two-level cache-aware approach for CPUs, we introduce a lightweight dynamic scheduling scheme. Before thread assignment, the scheme calculates the execution cost of each task for a given L3 chunk. Tasks are then stored in a shared queue, which is reconstructed per L3 chunk, and dispatched across CPU cores. The cost is defined as the exact number of non-masked tiles visited. Under a static causal mask, it can be derived directly from the query block index. We then sort the $q$-groups in descending order of cost and insert them into a lock-free shared queue. Instead of relying on static partitions, threads from different cores dynamically fetch the highest-cost remaining $q$-group using an atomic increment on a shared index. Threads that complete early immediately retrieve additional work from the queue, balancing the workload across all active cores and eliminating idle wait time.

\begin{equation}
\label{eq:Cost of a q-group}
\begin{aligned}
C = \sum_{n=0}^{G-1} \frac{\max\left(0,\ \min(L_n, L_{\text{end}}) - L_{\text{begin}}\right)}{T_{kv}}
\end{aligned}
\end{equation}

where $C$ is the cost of the $q$-group, $L_n$ refers to the longest sequence length in the $n$-th $q$-block while
$L_{\text{begin}}$ and $L_{\text{end}}$ denote the start and end indices of this chunk, $T_{kv}$ is the tile (step) size.

A key advantage of this design is that the dynamic queue naturally adapts to heterogeneous CPU architectures. Performance cores (P-cores) finish tasks faster and consequently pull a larger share of work from the queue, inherently mitigating load imbalance between P-cores and Efficient cores (E-cores) without requiring manual thread-to-core mapping.

\subsection{Combined Scheduling Methodology}
By combining the two proposed techniques described above, we establish an optimized execution scheme for causal attention computation. Algorithm \ref{alg:Tiered Attention on CPU} outlines the high-level structure of this approach. Unless explicitly defined, mathematical notations follow standard conventions in attention literature. For instance, $O$ denotes the output tensor and $B$ represents the batch size. For clarity, certain non-essential implementation details in the algorithm listing are simplified to provide a clear conceptual overview.
\begin{algorithm}[tb]
   \caption{Tiered Attention on CPU}
   \label{alg:Tiered Attention on CPU}
\begin{algorithmic}
   \STATE {\bfseries Input:} Tensors $Q$, $K$, $V$ in [B,H,S,D], scale
   \STATE {\bfseries Output:} Tensor $O$ in [B,H,S,D]
   \STATE Allocate global buffer: $O$, $m$, $l$
   \STATE Allocate per-thread buffer $S$, $P$
   \FOR [L3 cache-aware] {$hb=0$ {\bfseries to} $hb_\text{blocks}$}
   \STATE Pack K and V
   \FOR [L3 cache-aware]{$c=0$ {\bfseries to} $n_\text{chunk}$}
   \STATE Calculate the cost of various $q$-groups  
   \STATE Sort and put $q$-groups into an atomic queue
   \REPEAT[parfor - executed by each CPU thread]
   \STATE Fetch a $q$-group from the queue
   \FOR[L2 cache-aware] {$j = \text{chunk}_{start}$ \textbf{to} $\text{chunk}_{end}$}
   \FOR{$q_b = 0$ \textbf{to} $G - 1$}
    \IF {\text{Fully Masked}}
    \STATE $\text{Continue}$
    \ENDIF
   \STATE $S_b \gets scale*Q_bK_{j}^T$
   \STATE $m' \gets max(m, rowmax(S_b))$
   \STATE $P_b \gets exp2(S_b - m')$
   \STATE $l \gets 2^{(m-m')} * l + rowsum(P_b)$
   \STATE $O  \gets 2^{(m-m')} * O + P_b V_{j} $
   \STATE $m \gets m'$
   \ENDFOR
   \ENDFOR
   \UNTIL $\text{len(queue)} = 0$
   \ENDFOR
   \STATE Normalize $O \gets O/l$
   \ENDFOR
\end{algorithmic}
\end{algorithm}

\subsection{Overlapping GEMM and Softmax}
Inspired by the asynchronous execution model in FlashAttention-3~\citep{shah2024flashattention3}, we design an instruction-level pipeline that interleaves GEMM and Softmax computations on Intel Xeon processors. Specifically, we exploit instruction-level parallelism by offloading matrix multiplications ,$QK^T$, to Advanced Matrix Extensions (AMX) hardware accelerators, while concurrently executing online softmax vector operations on the prior score tile using AVX-512 SIMD. To be specific, step $j$ issues an AMX systolic GEMM for block $S_j=Q_b K_j^\top$ while AVX-512 vector units simultaneously process the exponentiation and reduction of $S_{j-1} = Q_b K_{j-1}^\top$. This fine-grained instruction overlap effectively hides softmax latency behind GEMM.

\begin{table*}[t]
\caption{Performance of different kernels on different sequence lengths using 1 batch size, 128 head dimensions}
\label{prefill-cacpu-comparison-table-selected}
\vskip 0.15in
\begin{center}
\begin{small}
\begin{sc}
\begin{tabular}{lcccccccr}
\toprule
Input ($S$) & CPU & GQA & SDPA (ms) & SGLANG (ms) & FLEX (ms) & Ours (ms) & Ratio \\
\midrule
4K & 6767P & 32:8 & 8.43 & 8.26 & 11.85 & \textbf{5.00} & 1.65\text{x} \\
8K & 6767P & 32:8 & 27.46 & 63.22 & 39.74 & \textbf{15.74} & 1.74\text{x} \\
16K & 6767P & 32:8 & 102.21 & 204.67 & 151.52 & \textbf{54.72} & 1.87\text{x} \\
32K & 6767P & 32:8 & 391.52 & 492.22 & 595.84 & \textbf{207.79} & 1.88\text{x} \\
64K & 6767P & 32:8 & 1562.45 & 1966.87 & 2339.86 & \textbf{812.10} & 1.92\text{x} \\
128K & 6767P & 32:8 & 6376.64 & 8147.56 & 9495.27 & \textbf{3170.23} & 2.01\text{x} \\
4K & 265K & 32:8 & 132.85 & N/A & 281.95 & \textbf{87.81} & 1.51\text{x} \\
8K & 265K & 32:8 & 503.39 & N/A & 1068.86 & \textbf{331.00} & 1.52\text{x} \\
16K & 265K & 32:8 & 1917.74 & N/A & 4397.27 & \textbf{1270.03} & 1.51\text{x} \\
32K & 265K & 32:8 & 7619.77 & N/A & 17634.20 & \textbf{4987.37} & 1.53\text{x} \\
64K & 265K & 32:8 & 31574.44 & N/A & 70803.12 & \textbf{19997.34} & 1.58\text{x} \\
128K & 265K & 32:8 & 126203.72 & N/A & 285030.40 & \textbf{79684.60} & 1.58\text{x} \\
\midrule
4K & 6767P & - & 8.56 & 8.23 & 12.24 & \textbf{4.70} & 1.75\text{x} \\
8K & 6767P & - & 28.13 & 56.41 & 39.91 & \textbf{17.12} & 1.64\text{x} \\
16K & 6767P & - & 107.02 & 150.20 & 152.62 & \textbf{56.41} & 1.90\text{x} \\
32K & 6767P & - & 407.19 & 499.66 & 601.17 & \textbf{219.62} & 1.85\text{x} \\
64K & 6767P & - & 1630.19 & 1942.53 & 2381.47 & \textbf{838.37} & 1.94\text{x} \\
128K & 6767P & - & 6596.03 & 7875.30 & 9711.11 & \textbf{3279.43} & 2.01\text{x} \\
4K & 265K & - & 135.84 & N/A & 295.22 & \textbf{93.40} & 1.45\text{x} \\
8K & 265K & - & 512.18 & N/A & 1109.84 & \textbf{346.42} & 1.48\text{x} \\
16K & 265K & - & 1958.13 & N/A & 4515.25 & \textbf{1321.08} & 1.48\text{x} \\
32K & 265K & - & 8013.11 & N/A & 18095.46 & \textbf{5274.21} & 1.52\text{x} \\
64K & 265K & - & 30621.06 & N/A & 72183.79 & \textbf{20668.62} & 1.48\text{x} \\
128K & 265K & - & 126458.67 & N/A & 290394.24 & \textbf{81932.76} & 1.54\text{x} \\
\bottomrule
\end{tabular}
\end{sc}
\end{small}
\end{center}
\vskip -0.1in
\end{table*}

\section{Evaluations}
We evaluate several aspects of the proposed kernel across two hardware platforms: a server-grade Intel Xeon 6767P (Xeon 6767P) and a desktop-grade Intel Core Ultra 7 265K (265K). Throughout the evaluations, K represents 1,024 when denoting sequence lengths and Grouped-Query Attention (GQA) 32:8 refers to 32 query heads sharing 8 key-value heads.

\subsection{Kernel Performance}
We evaluate and compare the performance of several attention kernels: PyTorch SDPA, PyTorch FlexAttention, SGLang, and the proposed TACPU kernel. Because the SGLang kernel requires hardware support for AMX, which is unavailable on the 265K processor, SGLang benchmarking was excluded for that CPU. Table \ref{prefill-cacpu-comparison-table-selected} summarizes the results, where the Ratio column highlights the latency speedup of the TACPU kernel relative to the best-performing alternative kernel for each test case.

As shown in Table \ref{prefill-cacpu-comparison-table-selected}, the TACPU kernel achieves the lowest latency. On the Xeon 6767P, TACPU is already $1.65\times$ to $1.75\times$ faster than the best baseline at 4K tokens, where K and V fit entirely in L3. The advantage grows to $2.01\times$ at 128K tokens. At longer sequences, where K and V exceed the L3 capacity, the L3-aware grouping combined with load-balanced scheduling sustains this advantage.

On the 265K, the advantage is roughly constant ($1.45$--$1.58\times$) across sequence lengths. 

\subsection{E2E Performance}
To evaluate the real-world performance of our proposed algorithm, we benchmark Time-to-First-Token (TTFT) prefill throughput within the SGLang serving framework against the best baseline, which is the PyTorch SDPA. Evaluations were conducted across Llama-3.1-8B, Qwen3-8B, and Mistral-7B-v0.3 over varying sequence lengths. Context lengths outside a model’s native support were evaluated and marked as 'timed-only'. Due to system memory constraints on the client-grade system (265K), tests for the 128K context length are omitted on that platform.

\begin{figure}[b!]
    \centering
    \includegraphics[scale=0.5]{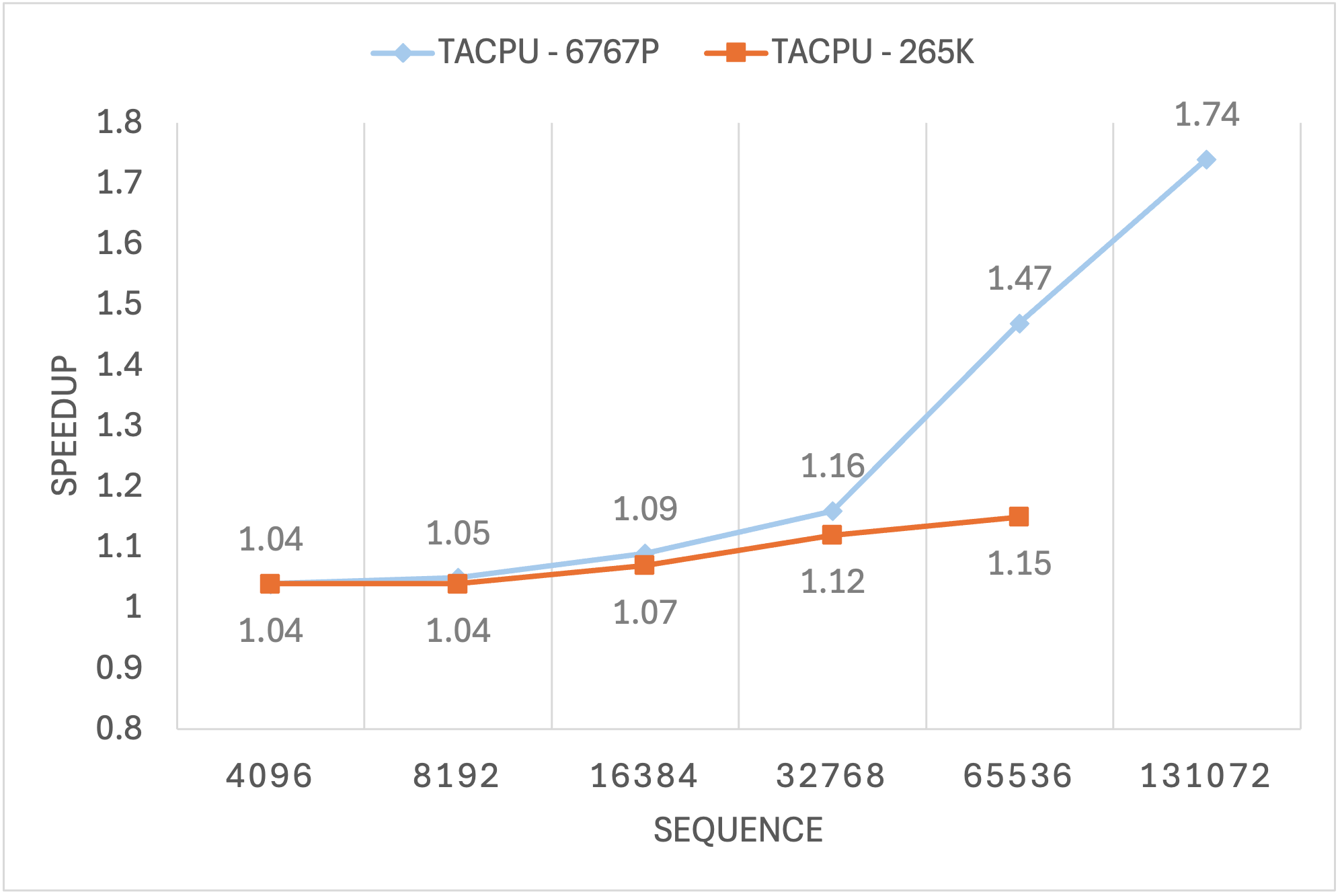}
    \caption{Prefill (TTFT) results of Llama-3.1-8B on Xeon 6767P and 265K}
    \label{fig:ttft_llama31_8b_fig}
\end{figure}

\begin{figure*}[t]
    \centering
    \begin{subfigure}
        \centering
        \includegraphics[scale=0.5]{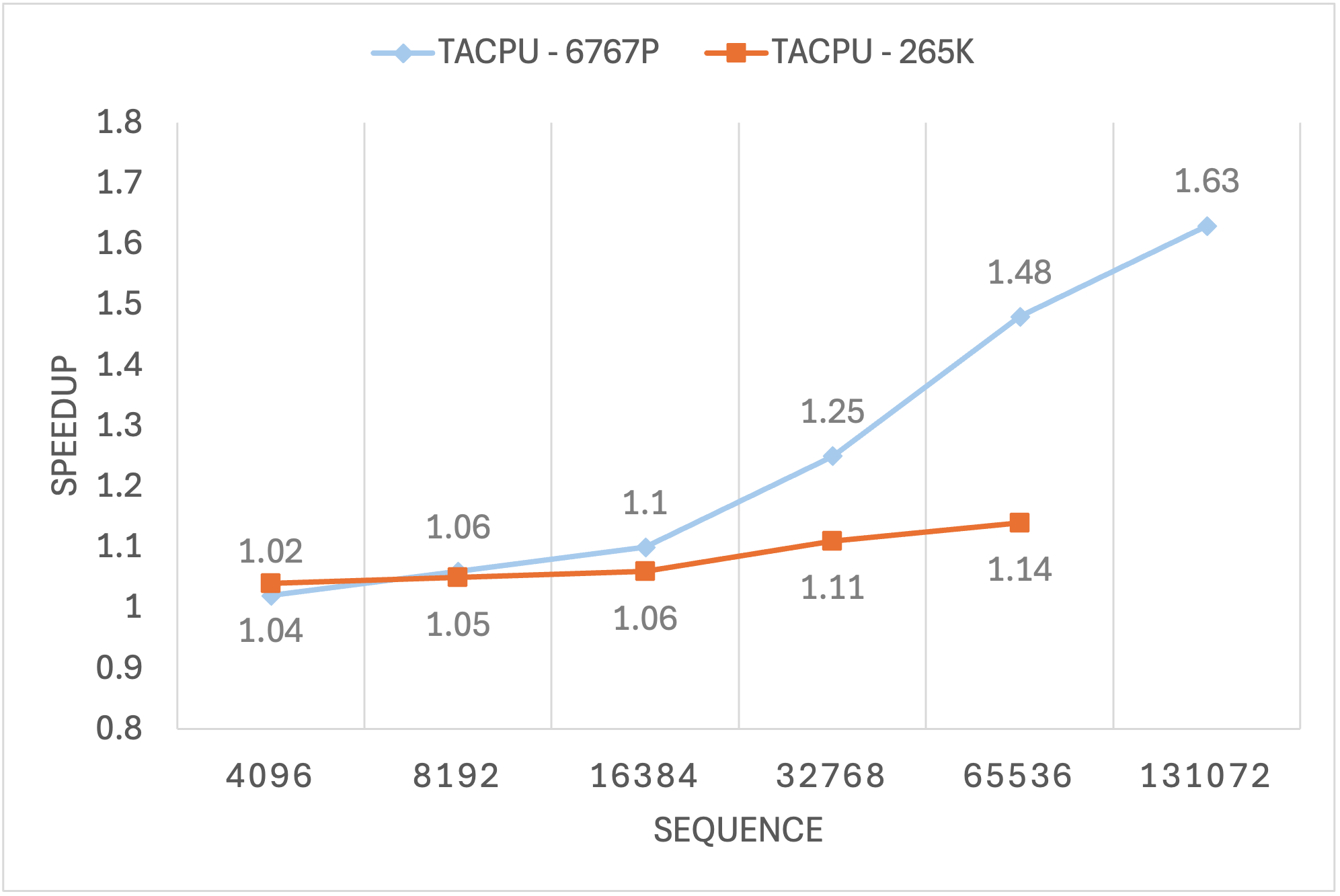}
        \label{fig:ttft_mistral_7b_v03_fig}
    \end{subfigure}
    \hfill 
    \begin{subfigure}
        \centering
        \includegraphics[scale=0.5]{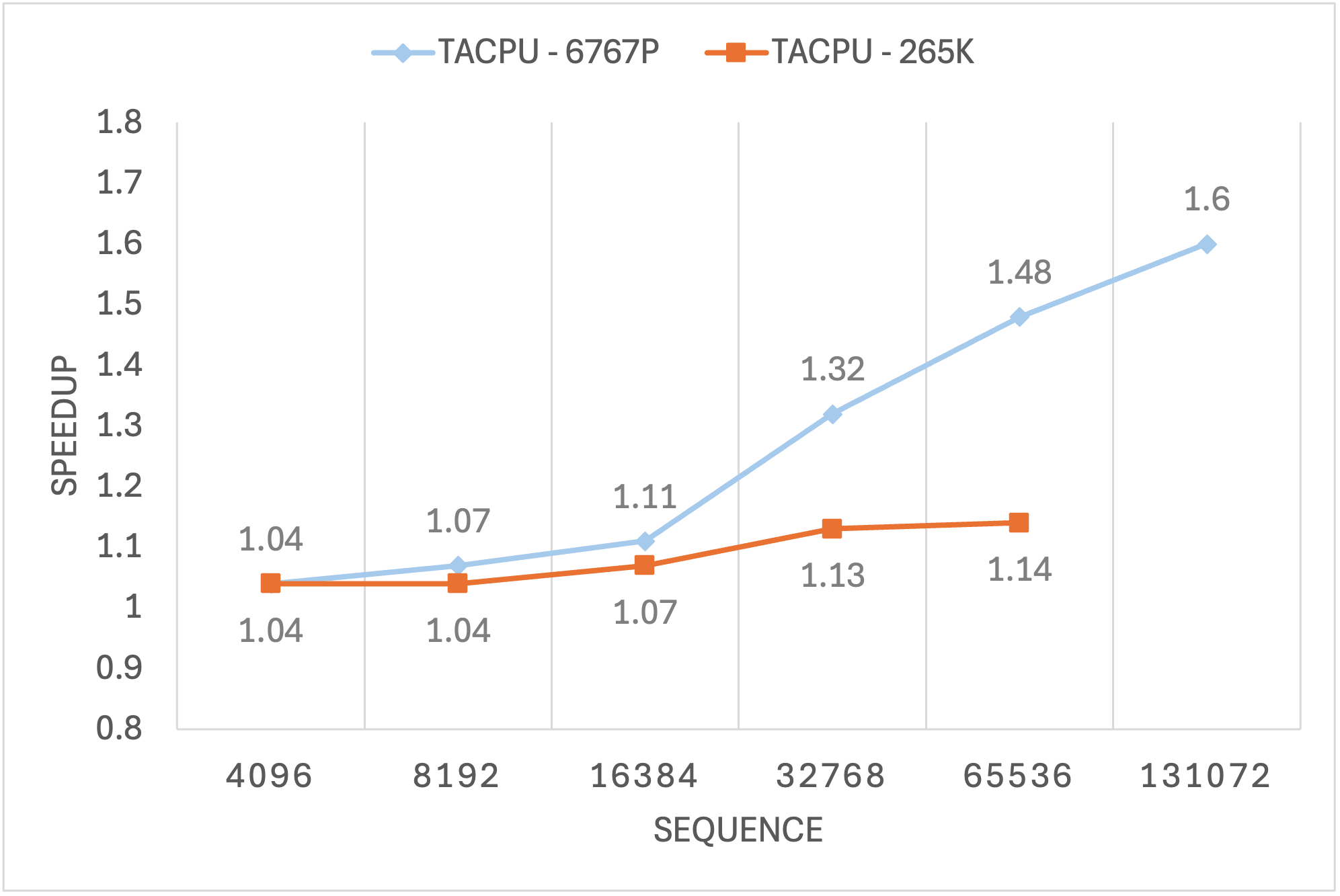}
        \label{fig:ttft_qwen3_8b_fig}
    \end{subfigure}
    \caption{Prefill (TTFT) results on Xeon 6767P and 265K. Left: Mistral-7B-v0.3 Right: Qwen3-8B}
    \label{fig:ttft_mistral_7b_v03_qwen3_8b_fig}
\end{figure*}

As shown in Figures \ref{fig:ttft_llama31_8b_fig}, and \ref{fig:ttft_mistral_7b_v03_qwen3_8b_fig}, our implementation achieves consistent throughput gains across both server-grade (Xeon 6767P) and client-grade (265K) processors. Speedup scales with sequence length, highlighting the efficiency of our algorithm on longer contexts. The performance improvement is particularly prominent on the Xeon 6767P, reaching up to a $\sim1.74\times$ speedup at a sequence length of 128K. This difference is due to the Xeon's native AMX acceleration for other GEMM operations, whereas the 265K lacks AMX support. Other non-attention GEMM calculations on 265K occupied significant time. As the sequence length grows, the proportion of computation time consumed by the attention mechanism increases. Consequently, the advantages of the proposed algorithm become evident in these scenarios. By addressing the attention bottleneck, our kernel significantly improves prefill throughput on AMX-enabled CPUs.

\FloatBarrier


\begin{figure*}[b!]
    \centering
    \begin{subfigure}
        \centering
        \includegraphics[scale=0.61]{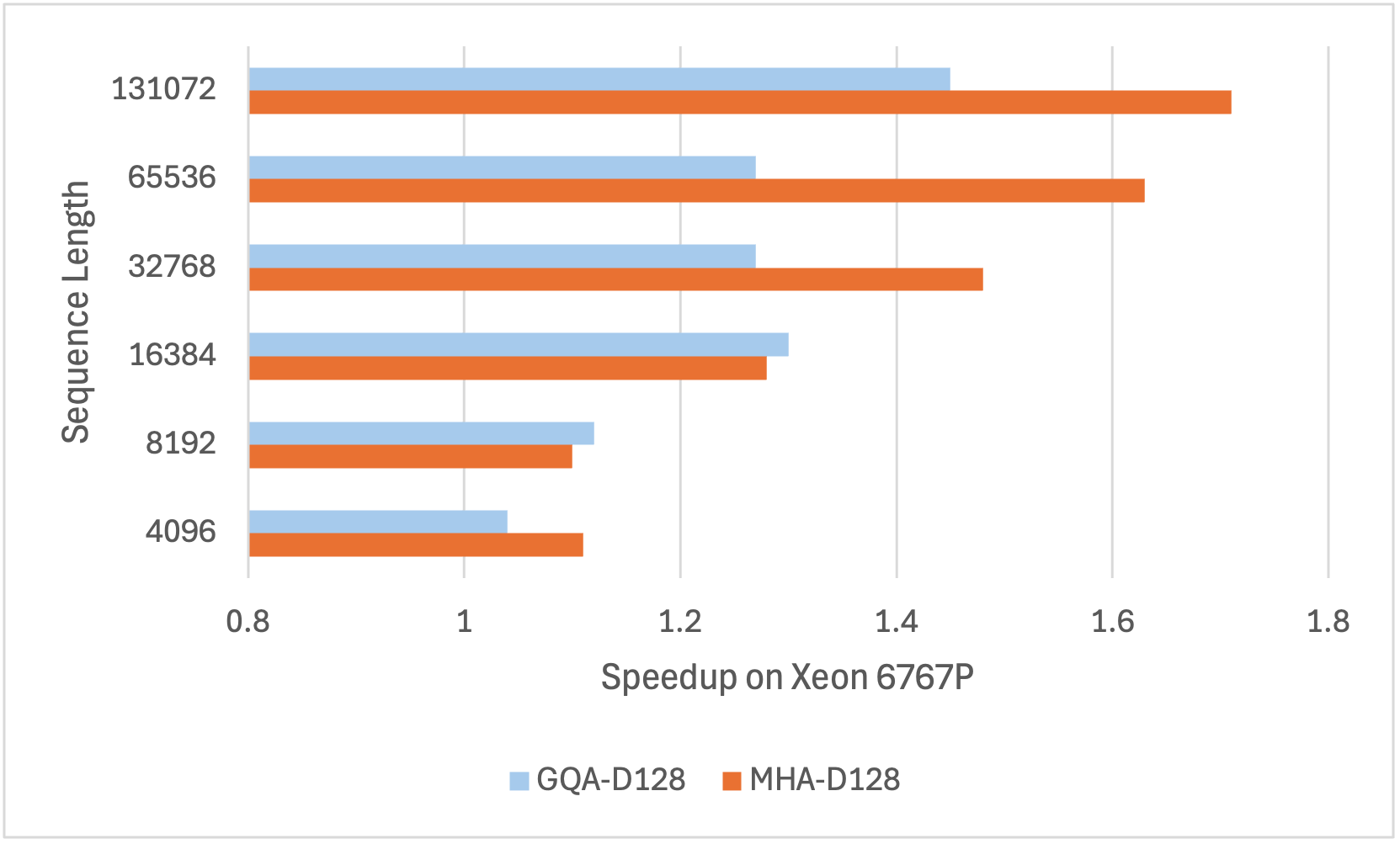}
        \label{fig:comparison_dynamic_scheduling_6767P_v2}
    \end{subfigure}
    \hfill 
    \begin{subfigure}
        \centering
        \includegraphics[scale=0.61]{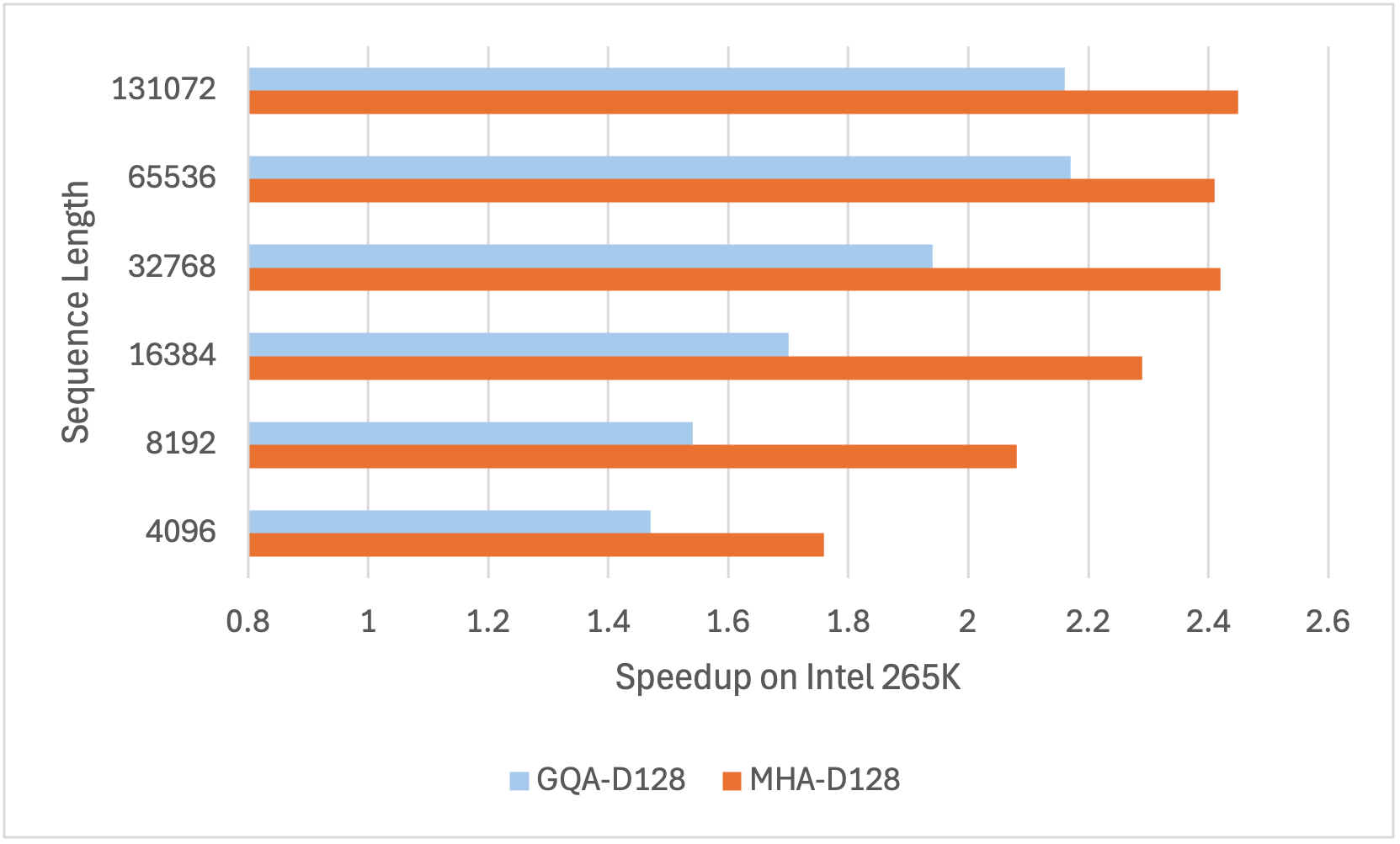}
        \label{fig:comparison_dynamic_scheduling_265k_v2}
    \end{subfigure}
    \caption{Comparative analysis of the L3-Chunk-Scoped Cost-Prioritized on Xeon 6767P (Left) and 265K (Right)}
    \label{fig:comparison_dynamic_scheduling_6767P_265k_v2}
\end{figure*}



\subsection{Analysis of L3-Chunk-Scoped Cost-Prioritized Scheduling}
To evaluate the efficacy of our proposed cost-prioritized scheduling scheme, we benchmarked its performance across various input shapes against the baseline static schedule, which is the original implementation without cost calculation or task sorting. We kept the other proposed techniques enabled. The results are presented in Figures \ref{fig:comparison_dynamic_scheduling_6767P_265k_v2}.

On a server-grade Xeon 6767P processor, the performance gains achieved by cost-prioritized scheduling scale proportionally with sequence length, delivering up to a $1.71\times$ speedup for long sequence lengths and Multi-Head Attention (MHA) configurations.

On a client-grade 265K processor, the impact of dynamic scheduling is even more pronounced. Performance improvements begin at short sequence lengths immediately, achieving a $1.72\times$ speedup. It then scales up to $2.45\times$ for longer sequences. This significant gain is caused by the scheme's ability to mitigate load imbalances inherent to heterogeneous architectures, which contain two distinct P-cores and E-cores. By utilizing a shared work queue, higher-throughput P-cores proactively retrieve and execute a larger portion of work items while waiting for E-cores to complete. Furthermore, because work items are sorted by the estimated cost, synchronization overhead in the final processing wave is significantly minimized.

The above results demonstrate that the proposed scheduling mechanism delivers substantial performance benefits across both high-performing server platforms and heterogeneous client processors.

\begin{table}[t]
\caption{Causal attention latency of the Cost-Prioritized and FIFO work-queue
schedules across different sequence lengths}
\label{tab:tiles-vs-fifo}
\vskip 0.15in
\begin{center}
\begin{small}
\begin{sc}
\setlength{\tabcolsep}{2pt} 
\begin{tabular}{lccccr}
\toprule
CPU & $S$ & FIFO (ms) & Ours (ms) & Speedup \\
\midrule
6767P & 4K   &    6.60 &    5.12 & \textbf{1.29}$\times$ \\
6767P & 8K   &   19.21 &   15.53 & \textbf{1.24}$\times$ \\
6767P & 16K  &   59.18 &   53.02 & \textbf{1.12}$\times$ \\
6767P & 32K  &  212.98 &  206.06 & \textbf{1.03}$\times$ \\
6767P & 64K  &  822.41 &  783.67 & \textbf{1.05}$\times$ \\
6767P & 128K & 3520.41 & 3119.72 & \textbf{1.13}$\times$ \\
\midrule
265K  & 4K   &    92.86 &    89.05 & \textbf{1.04}$\times$ \\
265K  & 8K   &   351.08 &   331.89 & \textbf{1.06}$\times$ \\
265K  & 16K  &  1341.50 &  1287.81 & \textbf{1.04}$\times$ \\
265K  & 32K  &  5366.40 &  5049.58 & \textbf{1.06}$\times$ \\
265K  & 64K  & 21128.56 & 20349.51 & \textbf{1.04}$\times$ \\
265K  & 128K & 82110.43 & 81328.25 & \textbf{1.01}$\times$ \\
\bottomrule
\end{tabular}
\end{sc}
\end{small}
\end{center}
\end{table}
\begin{table}[t]
\centering
\caption{Representative sequences demonstrating speedup under Cost-Prioritized Scheduling}
\label{tab:tiles-vs-index}
\vskip 0.15in
\begin{center}
\begin{small}
\begin{sc}
\begin{tabular}{lccccr}
\toprule
CPU & $S$ & Index(ms) & Ours (ms) & Speedup \\
\midrule
6767P & 1024 & 0.87 & 0.80 & \textbf{1.09}$\times$ \\
6767P & 4928 & 6.74 & 6.57 & \textbf{1.03}$\times$ \\
\midrule
265K & 1056 &  9.36 &  8.91 & \textbf{1.05}$\times$ \\
265K & 1824 & 21.19 & 20.86 & \textbf{1.02}$\times$ \\
\bottomrule
\end{tabular}
\end{sc}
\end{small}
\end{center}
\end{table}

We further compare First-In-First-Out (FIFO) scheduling with cost-prioritized scheduling. As shown in Table~\ref{tab:tiles-vs-fifo}, cost-prioritized scheduling achieves higher throughput on both CPUs, delivering an average speedup of 14.33\% on Xeon 6767P and 4.16\% on 265K. 

While relying on the $m$-block index \cite{zadouri2026flashattention4} yields comparable performance (within $\pm 1\%$) to our method in typical cases, Table~\ref{tab:tiles-vs-index} highlights scenarios where calculating the exact cost of each tile becomes critical. Performance degrades when a short tail $Q$-group is scheduled at the very end of the tile queue. Consequently, computing individual tile costs proves to be a more robust approach.

\begin{table}[t!]
\caption{DRAM read (GB) of different kernels across various sequence lengths}
\label{tab:DRAM-read}
\vskip 0.15in
\begin{center}
\begin{small}
\begin{sc}
\setlength{\tabcolsep}{4pt}
\begin{tabular}{lcccccr}
\toprule
CPU & $S$ & SDPA & SGLANG  & FLEX & Ours \\
\midrule
6767P & 4K   & 0.05    & 0.04   & 0.05    & \textbf{0.03}   \\
6767P & 8K   & 0.15    & 0.27   & 0.12    & \textbf{0.09}   \\
6767P & 16K  & 0.62    & 1.66   & 0.59    & \textbf{0.42}   \\
6767P & 32K  & 2.58    & 9.79   & 2.34    & \textbf{1.34}   \\
6767P & 64K  & 23.29   & 53.40  & 21.53   & \textbf{5.42}   \\
6767P & 128K & 150.50  & 397.77 & 210.53  & \textbf{21.13}  \\
\midrule
265K  & 4K   & 1.21    & N/A    & 0.40    & \textbf{0.12}   \\
265K  & 8K   & 5.73    & N/A    & 2.42    & \textbf{0.50}   \\
265K  & 16K  & 27.03   & N/A    & 22.95   & \textbf{3.17}   \\
265K  & 32K  & 110.43  & N/A    & 148.13  & \textbf{15.35}  \\
265K  & 64K  & 439.09  & N/A    & 768.34  & \textbf{108.75} \\
265K  & 128K & 1774.27 & N/A    & 3201.66 & \textbf{517.15} \\
\bottomrule
\end{tabular}
\end{sc}
\end{small}
\end{center}
\vskip -0.1in
\end{table}


\begin{table}[t!]
\caption{L3 Cache Hit Rate under various sequence lengths on 265K}
\label{tab:L3-Cache-Hit-Rate}
\vskip 0.15in
\begin{center}
\begin{small}
\begin{sc}
\begin{tabular}{lcccr}
\toprule
$S$ & SDPA(\%) & FLEX(\%) & Ours (\%)\\
\midrule
4K   & 84.42 & 94.30 & \textbf{98.68} \\
8K   & 79.78 & 90.28 & \textbf{97.48} \\
16K  & 78.18 & 69.37 & \textbf{96.50} \\
32K  & 74.92 & 52.62 &\textbf{ 94.66} \\
64K & 75.02 & 39.75 & \textbf{90.50} \\
128K & 75.67 & 33.78 & \textbf{88.29} \\
\bottomrule
\end{tabular}
\end{sc}
\end{small}
\end{center}
\vskip -0.1in
\end{table}



\subsection{Analysis of the Two-Level Cache-Aware Scheduling}
To evaluate the effect of cache-aware scheduling, we use Intel VTune Profiler alongside hardware Performance Monitoring Unit (PMU) counters to track two key metrics: DRAM read volume and L3 cache hit rate. We will also show the performance of the default $G$ value in $q$-block grouping.

\textbf{DRAM Read Volume}: We evaluated a representative shape ($[1, 32, S, 128]$ with a 32:8 GQA ratio) to demonstrate the effectiveness of L3-cache-aware scheduling. As shown in Table \ref{tab:DRAM-read}, TACPU significantly reduces DRAM reads across all sequence lengths on both CPU architectures. This reduction confirms that our scheduling strategy effectively eliminates redundant re-reads of Key and Value tensors from main memory.

\textbf{L3 Cache Hit Rate}: Benchmarks were conducted on the 265K processor, whose smaller L3 cache capacity exposes memory hierarchy limits at shorter sequence lengths. Note that SGLang was excluded from this comparison due to its lack of Intel Advanced Matrix Extensions (AMX) support on this processor architecture. As shown in Table \ref{tab:L3-Cache-Hit-Rate}, our method consistently outperforms both baseline kernels in L3 hit rate across all evaluated sequence lengths. By maintaining an average L3 hit rate of 94.35\%, TACPU demonstrates that its cache-aware design optimizes data locality for both short and long context windows.

\textbf{Grouping of $G$ $q$-blocks}: Using the same representative GQA configuration ($[1, 32, S, 128]$, 32:8 ratio), we evaluated how query-block ($q$-block) grouping affects latency. As reported in the Appendix, the default $G$ value yields competitive performance across sequence lengths. On average, the default $G$ setting is only 0.58\% slower than the optimal setting on the 265K and 2.34\% slower on the Xeon 6767P.

\subsection{Performance Compared to vLLM Attention Scheduling}
\begin{figure*}[t!]
    \centering
    \includegraphics[scale=0.62]
    {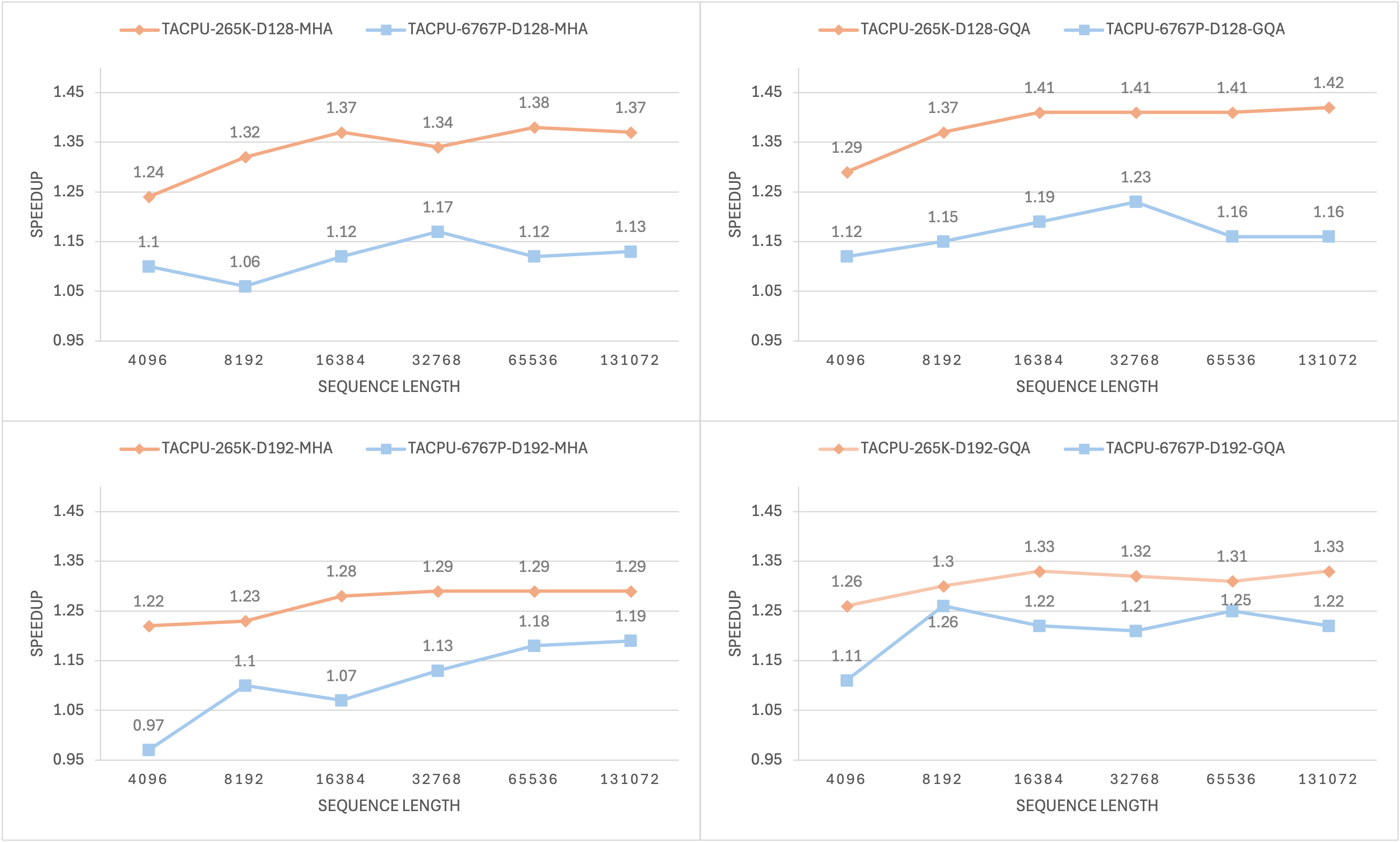}
    \caption{Comparative analysis of scheduling strategies in TACPU and vLLM. Top left: Results of Xeon 6767P and 265K with MHA and Head Dim. 128. Top right: Results of Xeon 6767P and 265K with GQA(32:8) and Head Dim. 128. Bottom left: Results of Xeon 6767P and 265K with MHA and Head Dim. 192. Bottom right: Results of Xeon 6767P and 265K with GQA(32:8) and Head Dim. 192}
    \label{fig:comparison_vllm_scheduling}
\end{figure*}

vLLM has a unique attention scheduler~\cite{li_2025_cpu_vllm}. We enabled this scheduler and compared the vLLM kernel with TACPU to check the effects. We set the batch size to 1 and used 32 query heads. Both multi-head attention (MHA) and grouped-query attention (GQA with a 32:8 query-to-KV ratio) were benchmarked. To evaluate performance under a more computationally demanding attention mechanism, we also conducted an experiment with a head dimension of 192. As shown in Figure \ref{fig:comparison_vllm_scheduling}, TACPU outperforms vLLM's backend in most cases. On the heterogeneous CPU (265K), the improvements are significant. These improvements are due to the proposed load-balanced scheduling, which minimizes synchronization time in the final wave and distributes workloads evenly through a shared queue. Because the proposed scheduler is integrated into the kernel itself, the benefits of the proposed method is transferable across different frameworks. No extra code has to be implemented outside.

\subsection{Forward Ablation Study}
To evaluate the cumulative impact of our proposed techniques, we conducted a forward ablation study on the attention kernel using a standard benchmark shape of $[1, 32, S, 128]$ with Grouped-Query Attention (GQA 32:8). Sequence lengths ($S$) were set to 64K and 128K on the 265K, and 128K and 256K on the Xeon 6767P, ensuring a sufficiently large memory footprint to trigger the L3 cache blocking optimization. Results are summarized in Table \ref{Ablations Study}.

As shown in Table \ref{Ablations Study}, enabling cost-prioritized scheduling first yields only a $1.08\times$ speedup on the Xeon processor at a 128K sequence length, whereas removing it last from the fully optimized settings results in a $1.45\times$ slowdown (Figure \ref{fig:comparison_dynamic_scheduling_6767P_265k_v2}). This discrepancy reveals a strong interaction effect. Without the optimizations, execution is limited by memory stalls. Cache-aware blocking eliminates this memory bottleneck, shifting the limiting factor to thread imbalance.

On the client-side 265K CPU, the performance divergence between P-cores and E-cores is the major limiting factor. Enabling load-balanced scheduling immediately yields a $1.44\times$ latency improvement. By comparing the results with Figure \ref{fig:comparison_dynamic_scheduling_6767P_265k_v2}, we find that the Two-Level Cache-Aware Scheduling is not an independent optimization. Enabling cache-aware scheduling by itself degrades performance, which explains why enabling the cost-prioritized scheduling at last brings significant gains (2.45$\times$). Two-Level Cache-Aware Scheduling only improves latency when integrated with the cost-prioritized scheduling, as the cache blocking mechanism aggravates load imbalances across P and E cores.

These findings demonstrate that cache-aware and cost-prioritized scheduling are complementary. These optimizations act synergistically to reduce overall latency.



\begin{table*}[t!]
\caption{Forward ablation study of the proposed algorithm across long sequences}
\label{Ablations Study}
\vskip 0.15in
\begin{center}
\begin{small}
\begin{sc}
\setlength{\tabcolsep}{6pt} 
\begin{tabular}{lccccr}
\toprule
CPU & $S$ & Technique & Latency(ms) & Speedup versus last step\\
\midrule
6767P & 128K & All Off & 8512.65 & - \\
6767P & 128K & + L3-Chunk-Scoped Cost-prioritized Scheduling & 7883.39 & 1.08$\times$ \\
6767P & 128K & + L3 Blocking (Two-Level Cache-Aware Scheduling) & 5884.57 & 1.34$\times$ \\
6767P & 128K & + GEMM and Softmax Overlap & 5239.27 & 1.12$\times$ \\
6767P & 128K & + Q-Grouping (Two-Level Cache-Aware Scheduling) & 3147.51 & 1.66$\times$ \\
\midrule
6767P & 256K & All Off & 30372.22 & - \\
6767P & 256K & + L3-Chunk-Scoped Cost-prioritized Scheduling & 28790.63 & 1.05$\times$ \\
6767P & 256K & + L3 Blocking (Two-Level Cache-Aware Scheduling) & 23044.99 & 1.25$\times$ \\
6767P & 256K & + GEMM and Softmax Overlap & 20930.11 & 1.10$\times$ \\
6767P & 256K & + Q-Grouping (Two-Level Cache-Aware Scheduling) & 12514.80 & 1.67$\times$ \\


\midrule
265K & 64K & All Off & 31514.57 & - \\
265K & 64K & + L3-Chunk-Scoped Cost-prioritized Scheduling & 21919.76 & 1.44$\times$ \\
265K & 64K & + L3 Blocking (Two-Level Cache-Aware Scheduling) & 22115.17 & 0.99$\times$ \\
265K & 64K & + Q-Grouping (Two-Level Cache-Aware Scheduling) & 20519.54 & 1.08$\times$ \\
\midrule
265K & 128K & All Off & 126582.79 & - \\
265K & 128K & + L3-Chunk-Scoped Cost-prioritized Scheduling & 87411.98 & 1.44$\times$ \\
265K & 128K & + L3 Blocking (Two-Level Cache-Aware Scheduling) & 87808.91 & 1.00$\times$ \\
265K & 128K & + Q-Grouping (Two-Level Cache-Aware Scheduling) & 81901.38 & 1.07$\times$ \\
\bottomrule
\end{tabular}
\end{sc}
\end{small}
\end{center}
\vskip -0.1in
\end{table*}

\subsection{Stability}
We verified the numerical correctness and stability of our kernel against FP64 PyTorch's eager mode scaled\_dot\_product\_attention function across all tested sequence lengths. The minimum cosine similarity achieved was 0.999, and the maximum absolute difference was 0.01 across all the mentioned input shapes. To evaluate end-to-end model numerical stability, we integrated our custom TACPU kernel into SGLang and validated the integrated system on Llama-3.1-8B using two standard benchmarks, MMLU and RULER. Our kernel achieved an MMLU accuracy of 66.2\%, virtually identical to the official baseline (66.7\%). Similarly, RULER evaluations demonstrated consistent performance compared to the reference SDPA implementation.

\section{Conclusion}
In this work, we introduce CPU-centric optimizations to FlashAttention by incorporating cache-aware and cost-prioritized scheduling mechanisms. With an aim to minimize redundant DRAM reads, the algorithm first partitions long-sequence Key and Value tensors into chunks that fit within the L3 cache. At the L2 cache level, query-blocks are grouped to maximize reuse of the K and V subblocks residing in L2, significantly improving latency. To validate correctness, we verified the numerical differences at the single-kernel level using cosine similarity against reference outputs. We also evaluated the kernel within multi-layer models using the MMLU and RULER benchmarks, confirming that model accuracy and system robustness are fully preserved in end-to-end scenarios.

Extensive evaluations have been conducted to demonstrate the efficiency of the proposed kernel. In microbenchmarks, our implementation consistently outperformed standard baselines, which are PyTorch SDPA, SGLang's attention kernel, and PyTorch FlexAttention, by up to a $2.01\times$ speedup on long sequences. End-to-end evaluation using SGLang as the serving framework demonstrated corresponding improvements in Time-to-First-Token (TTFT) across multiple models (Llama-3.1-8B, Qwen3-8B, and Mistral-7B-v0.3), revealing the kernel's effectiveness in accelerating the LLM prefill phase. Furthermore, comparative analysis against vLLM’s framework-level attention scheduler paired with its optimized kernel shows that integrating scheduling directly inside the kernel achieved performance gains, while maintaining framework portability without requiring custom framework modifications. Finally, the experimental results confirm the contributions of both proposed mechanisms. Cache-aware scheduling improves L3 cache hit rates and reduces DRAM read volume, while cost-prioritized scheduling delivers up to a $1.44\times$ speedup on a heterogeneous CPU architecture. Both optimizations interact to substantially improve latency performance.

While the current algorithm effectively covers common scenarios, there are certain limitations. The current algorithm assumes standard causal attention. Modified attention patterns, such as attention sinks or custom sparsity masks, change the computational cost model used for load balancing and the memory access patterns. These can degrade the L2 and L3 cache locality. However, the underlying concepts are flexible. The scheduling criteria can be extended to accommodate prefill variants, custom masking patterns, and also decode. As the attention mechanisms evolve to support emerging LLM serving and agentic AI workloads, we will continue to improve this algorithm to support non-standard masking structures and variants of attention based on the principles proposed.

\bibliography{references_checked}
\bibliographystyle{mlsys2025}

\section{AI Usage Disclosure}
During the preparation of this paper, we used Gemini 3 and Opus 5/5.5 to refine the clarity, grammar, and format of initial human-authored drafts. Furthermore, Opus 5/5.5 was used to assist with programming tasks, such as structuring and drafting code for the kernel and evaluation scripts. Technical design decisions were performed by us.

\appendix
\clearpage

\section{Supplementary Experimental Results}
\subsection{Kernel Performance}
Table~\ref{tab:prefill-cacpu-comparison-table} compares kernel performance across various configurations. To show how performance scales with sequence length, all tests are conducted with a fixed batch size of 1.

\subsection{Effect of Grouping Parameter $G$}
To examine the impact of the grouping parameter $G$ on mean latency, we vary $G$ while holding all other parameters constant. Experiments are conducted using an input shape of $[1, 32, S, 128]$ under a $32:8$ GQA setting. The results are summarized in Table~\ref{tab:gsweep_265k} and Table~\ref{tab:gsweep_6767P}.

\subsection{TFLOPS Ultilizations}
We evaluate kernel TFLOPS utilization using a representative input shape of $[1, 32, S, 128]$ under a $32:8$ GQA configuration. Peak hardware throughputs for 265K and Xeon 6767P platforms are measured using a register-resident loop. The peak TFLOPs are 3.0 TFLOPS and 158.9 TFLOPS for 265K and Xeon 6767P respectively. Tables~\ref{tab:roofline-6767p} and ~\ref{tab:roofline-265K} shows the details. The TFLOP ultilization(\%) is calculated by using the unrounded TFLOPs, not the printed TFLOPs.

\section{Evaluation  Methodology}
In this section, we present the experimental setup used to evaluate the proposed algorithm. First, we outline the hardware platform and software environment. We then describe the benchmark input configurations and sequence shapes. Finally, we detail the kernel implementation details and outline our experimental evaluation methodology.

\subsection{Hardware Infrastructure}To evaluate both the execution performance and numerical accuracy of the proposed algorithm, we conduct experiments across two distinct CPU architectures:
\begin{itemize}
    \item \textbf{Enterprise Platform:} We use Intel Xeon 6767P (Xeon 6767P) as it supports Intel Advanced Matrix Extensions (AMX) and AVX-512 instruction sets, which enable fused and overlapped execution of GEMM and Softmax operators. It also features a large L3 cache (336MB).
    \item \textbf{Client Platform:} Intel Core Ultra 7 265K (265K) is used as it is a hybrid processor architecture (comprising P-cores and E-cores), which supports  AVX2 and is equipped with a relatively small 30 MB L3 cache.
    
\end{itemize}
Evaluating across these two platforms demonstrates how our cache-aware scheduling operates under various cache capacities and core topologies. Table \ref{tab:hardware} shows platform details.

\subsection{Software Infrastructure}
Table \ref{tab:hardware} also lists the versions of the primary software packages and runtimes used in our evaluation.

\subsection{Input and Benchmarks}
All benchmarks are evaluated using Brain Floating Point 16 (\texttt{bf16}), one primary numerical precision used in modern Large Language Model (LLM) inference. Inputs are formatted as 4D tensors with dimensions $[B, H, S, D]$. 
The kernel microbenchmarks evaluate full-sequence causal prefill, where the query and key/value tensors have equal sequence lengths. Most experiments use a batch size of 1 to isolate single-sequence kernel performance and examine its scaling with sequence length. We also evaluate across representative Multi-Head Attention (MHA, 32:32) and Grouped-Query Attention (GQA, 32:8) configurations. Sequence lengths ($S$) range from 4,096 (4K) to 131,072 (128K) tokens, with head dimensions ($D$) set to 128 or 192. If not specified, the input shape will be [1,32,$S$,128] with GQA 32:8 as this is the typical shape for models. For performance benchmarking, we adopted the following execution scheme: experiments on sequence lengths from 4K to 32K were measured over 20 iterations, while sequences of 64K and 128K were evaluated over 10 and 5 iterations respectively. The average latency across iterations is reported.

\subsection{Kernels}
We implement the proposed algorithm using the PyTorch Codegen mechanism. The custom kernel is authored as a C++ template, which is similar to the current existing CPU FlexAttention,  under PyTorch Codegen. This allows PyTorch to serve as a unified benchmarking runtime.  We selected the mainstream attention kernel implementations in SGLang, vLLM and PyTorch as our primary baselines, as they represent the current state of the art for LLM serving. The kernels extracted from SGLang and vLLM are built as custom operators and bound via the PyTorch TORCH\_LIBRARY mechanism. The standard baselines, which are the  torch.nn.functional.scaled\_dot\_product\_attention (SDPA) and torch.nn.attention.flex\_attention, are executed directly through the native PyTorch APIs without modification. We omit the CPU kernels from llama.cpp as baselines because they lack support for Intel AMX instructions. Initial benchmarking on our Xeon host demonstrated that llama.cpp is significantly slower.

\subsection{Kernel Parameters}
The parameters of the kernel used throughout the evaluations are listed in Table \ref{tab:kernel-params}. 

\begin{table*}[b!]
\caption{Performance of different kernels on different sequence lengths, head dimensions, and GQA settings}
\label{tab:prefill-cacpu-comparison-table}
\vskip 0.15in
\begin{center}
\begin{small}
\begin{sc}
\begin{tabular}{lcccccccr}
\toprule
Input ($S$,$D$) & CPU & GQA & SDPA (ms) & SGLANG (ms) & FLEX (ms) & TACPU (ms) & Ratio \\
\midrule
4096,128 & 6767P & - & 8.56 & 8.23 & 12.24 & \textbf{4.70} & 1.75\text{x} \\
8192,128 & 6767P & - & 28.13 & 56.41 & 39.91 & \textbf{17.12} & 1.64\text{x} \\
16384,128 & 6767P & - & 107.02 & 150.20 & 152.62 & \textbf{56.41} & 1.90\text{x} \\
32768,128 & 6767P & - & 407.19 & 499.66 & 601.17 & \textbf{219.62} & 1.85\text{x} \\
65536,128 & 6767P & - & 1630.19 & 1942.53 & 2381.47 & \textbf{838.37} & 1.94\text{x} \\
131072,128 & 6767P & - & 6596.03 & 7875.30 & 9711.11 & \textbf{3279.43} & 2.01\text{x} \\
4096,128 & 6767P & 32:8 & 8.43 & 8.26 & 11.85 & \textbf{5.00} & 1.65\text{x} \\
8192,128 & 6767P & 32:8 & 27.46 & 63.22 & 39.74 & \textbf{15.74} & 1.74\text{x} \\
16384,128 & 6767P & 32:8 & 102.21 & 204.67 & 151.52 & \textbf{54.72} & 1.87\text{x} \\
32768,128 & 6767P & 32:8 & 391.52 & 492.22 & 595.84 & \textbf{207.79} & 1.88\text{x} \\
65536,128 & 6767P & 32:8 & 1562.45 & 1966.87 & 2339.86 & \textbf{812.10} & 1.92\text{x} \\
131072,128 & 6767P & 32:8 & 6376.64 & 8147.56 & 9495.27 & \textbf{3170.23} & 2.01\text{x} \\
\midrule
4096,128 & 265K & - & 135.84 & N/A & 295.22 & \textbf{93.40} & 1.45\text{x} \\
8192,128 & 265K & - & 512.18 & N/A & 1109.84 & \textbf{346.42} & 1.48\text{x} \\
16384,128 & 265K & - & 1958.13 & N/A & 4515.25 & \textbf{1321.08} & 1.48\text{x} \\
32768,128 & 265K & - & 8013.11 & N/A & 18095.46 & \textbf{5274.21} & 1.52\text{x} \\
65536,128 & 265K & - & 30621.06 & N/A & 72183.79 & \textbf{20668.62} & 1.48\text{x} \\
131072,128 & 265K & - & 126458.67 & N/A & 290394.24 & \textbf{81932.76} & 1.54\text{x} \\
4096,128 & 265K & 32:8 & 132.85 & N/A & 281.95 & \textbf{87.81} & 1.51\text{x} \\
8192,128 & 265K & 32:8 & 503.39 & N/A & 1068.86 & \textbf{331.00} & 1.52\text{x} \\
16384,128 & 265K & 32:8 & 1917.74 & N/A & 4397.27 & \textbf{1270.03} & 1.51\text{x} \\
32768,128 & 265K & 32:8 & 7619.77 & N/A & 17634.20 & \textbf{4987.37} & 1.53\text{x} \\
65536,128 & 265K & 32:8 & 31574.44 & N/A & 70803.12 & \textbf{19997.34} & 1.58\text{x} \\
131072,128 & 265K & 32:8 & 126203.72 & N/A & 285030.40 & \textbf{79684.60} & 1.58\text{x} \\
\midrule[\heavyrulewidth]
4096,192 & 6767P & - & 10.22 & 12.18 & 12.88 & \textbf{7.66} & 1.33\text{x} \\
8192,192 & 6767P & - & 35.12 & 41.84 & 45.12 & \textbf{22.49} & 1.56\text{x} \\
16384,192 & 6767P & - & 137.17 & 185.92 & 175.75 & \textbf{77.87} & 1.76\text{x} \\
32768,192 & 6767P & - & 511.14 & 616.79 & 722.60 & \textbf{287.67} & 1.78\text{x} \\
65536,192 & 6767P & - & 2055.91 & 2415.13 & 2925.59 & \textbf{1087.57} & 1.89\text{x} \\
131072,192 & 6767P & - & 8421.99 & 9741.30 & 12581.35 & \textbf{4315.57} & 1.95\text{x} \\
4096,192 & 6767P & 32:8 & 10.08 & 12.40 & 12.81 & \textbf{6.59} & 1.53\text{x} \\
8192,192 & 6767P & 32:8 & 33.95 & 63.02 & 44.48 & \textbf{19.54} & 1.74\text{x} \\
16384,192 & 6767P & 32:8 & 123.44 & 264.93 & 170.48 & \textbf{69.33} & 1.78\text{x} \\
32768,192 & 6767P & 32:8 & 480.84 & 615.03 & 673.00 & \textbf{263.74} & 1.82\text{x} \\
65536,192 & 6767P & 32:8 & 1946.26 & 2442.70 & 2779.70 & \textbf{1041.74} & 1.87\text{x} \\
131072,192 & 6767P & 32:8 & 7836.07 & 10075.43 & 11645.74 & \textbf{4113.04} & 1.91\text{x} \\
\midrule
4096,192 & 265K & - & 182.54 & N/A & 401.25 & \textbf{132.54} & 1.38\text{x} \\
8192,192 & 265K & - & 693.02 & N/A & 1548.61 & \textbf{512.69} & 1.35\text{x} \\
16384,192 & 265K & - & 2647.73 & N/A & 6190.52 & \textbf{1982.49} & 1.34\text{x} \\
32768,192 & 265K & - & 11184.75 & N/A & 25075.55 & \textbf{7882.53} & 1.42\text{x} \\
65536,192 & 265K & - & 41453.36 & N/A & 98865.72 & \textbf{31646.59} & 1.31\text{x} \\
131072,192 & 265K & - & 164391.34 & N/A & 401299.55 & \textbf{125974.93} & 1.30\text{x} \\
4096,192 & 265K & 32:8 & 179.92 & N/A & 390.46 & \textbf{126.61} & 1.42\text{x} \\
8192,192 & 265K & 32:8 & 683.12 & N/A & 1515.38 & \textbf{481.51} & 1.42\text{x} \\
16384,192 & 265K & 32:8 & 2714.13 & N/A & 6099.51 & \textbf{1880.79} & 1.44\text{x} \\
32768,192 & 265K & 32:8 & 10399.19 & N/A & 24386.55 & \textbf{7638.37} & 1.36\text{x} \\
65536,192 & 265K & 32:8 & 43995.51 & N/A & 97933.12 & \textbf{30750.44} & 1.43\text{x} \\
131072,192 & 265K & 32:8 & 164950.04 & N/A & 392480.33 & \textbf{123192.83} & 1.34\text{x} \\
\bottomrule
\end{tabular}
\end{sc}
\end{small}
\end{center}
\vskip -0.1in
\end{table*}
\begin{table*}[b!]
\caption{Mean latency (ms) of TACPU on the Intel Core Ultra 7 265K across different grouping values}
\label{tab:gsweep_265k}
\vskip 0.15in
\begin{center}
\begin{small}
\begin{sc}
\begin{tabular}{lccccccr}
\toprule
$G$ & 4K (ms) & 8K (ms) & 16K (ms) & 32K (ms) & 64K (ms) & 128K (ms) \\
\midrule
1             & 89.5          & 336.9          & 1322.1          & 5312.6          & 21726.2          & 87370.0          \\
2             & \textbf{88.9} & 333.8          & 1306.3          & 5132.9          & 20446.6          & 81665.9          \\
4 (default)   & 89.7          & \textbf{333.6} & 1289.4          & 5093.8          & 20305.1          & 80543.8          \\
6             & 89.0          & 339.9          & 1284.4          & 5067.1          & 20214.9          & \textbf{80501.8} \\
8             & 89.6          & 334.3          & \textbf{1277.8} & 5061.8          & \textbf{20175.5} & 80563.4          \\
10            & 89.8          & 336.7          & 1288.4          & \textbf{5045.9} & 20393.8          & 80641.7          \\
\bottomrule
\end{tabular}
\end{sc}
\end{small}
\end{center}
\end{table*}
\begin{table*}[b!]
\caption{Mean latency (ms) of TACPU on the Xeon 6767P across different grouping values}
\label{tab:gsweep_6767P}
\vskip 0.15in
\begin{center}
\begin{small}
\begin{sc}
\begin{tabular}{lccccccr}
\toprule
$G$ & 4K (ms) & 8K (ms) & 16K (ms) & 32K (ms) & 64K (ms) & 128K (ms) \\
\midrule
1 & 6.44 & 19.85 & 74.40 & 271.60 & 1042.94 & 4947.31 \\
2 & 5.45 & 17.57 & 59.38 & 219.10 & 846.81 & 3361.34 \\
4 & 5.47 & 16.13 & 54.87 & 202.14 & 781.92 & 3105.14 \\
6 & 5.06 & 16.03 & 53.84 & 200.77 & \textbf{771.36} & \textbf{3090.10} \\
8 & 5.17 & 15.58 & 53.60 & 200.13 & 779.52 & 3135.92 \\
10 & 5.04 & 15.47 & 53.93 & 200.70 & 781.33 & 3153.19 \\
12 & \textbf{4.81} & 15.33 & 53.59 & 199.30 & 780.45 & 3152.97 \\
14 & 5.17 & 15.04 & 53.99 & 198.81 & 777.46 & 3141.97 \\
16 & 5.09 & 14.74 & 53.66 & \textbf{198.39} & 774.68 & 3147.21 \\
18 (default) & 5.12 & 14.99 & 53.79 & 198.80 & 779.08 & 3126.17 \\
20 & 5.21 & 14.72 & 53.04 & 198.66 & 775.45 & 3123.77 \\
22 & 5.46 & 14.65 & 53.21 & 199.13 & 777.72 & 3246.87 \\
24 & 5.31 & 15.18 & 53.11 & 199.28 & 778.61 & 3239.89 \\
26 & 5.25 & 14.82 & 53.27 & 201.32 & 891.70 & 3113.52 \\
28 & 5.26 & 15.19 & 53.03 & 201.88 & 912.57 & 3115.25 \\
30 & 5.21 & 14.94 & \textbf{52.88} & 202.51 & 910.53 & 3144.64 \\
32 & 5.32 & \textbf{14.48} & 53.15 & 202.01 & 778.40 & 3148.45 \\
34 & 5.28 & 16.19 & 53.73 & 203.23 & 909.87 & 3146.87 \\
36 & 5.27 & 15.17 & 54.09 & 201.83 & 909.79 & 3137.57 \\
38 & 5.33 & 15.83 & 54.09 & 202.13 & 907.06 & 3142.86 \\
40 & 5.28 & 15.42 & 53.85 & 200.50 & 796.31 & 3144.49 \\
42 & 5.27 & 15.79 & 54.18 & 201.18 & 931.32 & 3145.02 \\
44 & 5.30 & 15.78 & 54.04 & 201.42 & 900.51 & 3143.96 \\
46 & 5.32 & 15.75 & 55.18 & 202.42 & 890.07 & 3137.56 \\
48 & 5.32 & 15.35 & 55.36 & 200.02 & 788.51 & 3143.84 \\
50 & 5.25 & 15.76 & 54.00 & 202.20 & 909.47 & 3152.78 \\
\bottomrule
\end{tabular}
\end{sc}
\end{small}
\end{center}
\end{table*}
\begin{table*}[t]
\caption{TFLOPS utilization of different kernels across varying sequence lengths on the Xeon 6767P}
\label{tab:roofline-6767p}
\vskip 0.15in
\begin{center}
\begin{small}
\begin{sc}
\begin{tabular}{lcccccr}
\toprule
$S$ & Algorithm & TFLOP/s & TFLOP Util.\ (\%) & DRAM read (GB) & FLOP/B \\
\midrule
4K & SDPA & 14.01 & 8.82 & 0.05 & 2,923.40 \\
4K & SGLang & 16.99 & 10.69 & 0.04 & 3,523.08 \\
4K & Flex & 11.98 & 7.54 & 0.04 & 3,053.33 \\
4K & Ours & 29.33 & \textbf{18.46} & 0.03 & 4,432.26 \\
\midrule
8K & SDPA & 20.85 & 13.12 & 0.15 & 3,740.14 \\
8K & SGLang & 15.19 & 9.56 & 0.27 & 2,051.49 \\
8K & Flex & 14.47 & 9.11 & 0.12 & 4,620.17 \\
8K & Ours & 36.46 & \textbf{22.94} & 0.09 & 5,911.83 \\
\midrule
16K & SDPA & 22.43 & 14.11 & 0.62 & 3,546.77 \\
16K & SGLang & 15.49 & 9.75 & 1.66 & 1,327.90 \\
16K & Flex & 14.99 & 9.43 & 0.59 & 3,702.02 \\
16K & Ours & 42.06 & \textbf{26.47} & 0.41 & 5,298.80 \\
\midrule
32K & SDPA & 23.23 & 14.62 & 2.58 & 3,404.06 \\
32K & SGLang & 17.70 & 11.14 & 9.79 & 898.29 \\
32K & Flex & 15.02 & 9.45 & 2.34 & 3,762.23 \\
32K & Ours & 45.25 & \textbf{28.47} & 1.34 & 6,578.98 \\
\midrule
64K & SDPA & 22.56 & 14.20 & 23.29 & 1,510.84 \\
64K & SGLang & 15.88 & 9.99 & 53.40 & 658.87 \\
64K & Flex & 15.06 & 9.48 & 21.53 & 1,634.13 \\
64K & Ours & 45.95 & \textbf{28.92} & 5.42 & 6,485.60 \\
\midrule
128K & SDPA & 23.45 & 14.76 & 150.50 & 935.15 \\
128K & SGLang & 17.32 & 10.90 & 397.77 & 353.82 \\
128K & Flex & 15.48 & 9.74 & 210.53 & 668.50 \\
128K & Ours & 45.91 & \textbf{28.89} & 21.13 & 6,661.50 \\
\bottomrule
\end{tabular}
\end{sc}
\end{small}
\end{center}
\vskip -0.1in
\end{table*}
\begin{table*}[t]
\caption{TFLOPS utilization of different kernels across varying sequence lengths on the 265K CPU}
\label{tab:roofline-265K}
\vskip 0.15in
\begin{center}
\begin{small}
\begin{sc}
\begin{tabular}{lcccccr}
\toprule
$S$ & Algorithm & TFLOP/s & TFLOP Util.\ (\%) & DRAM read (GB) & FLOP/B \\
\midrule
4K & SDPA & 1.03 & 34.49 & 1.21 & 113.28 \\
4K & SGLang & N/A & N/A & N/A & N/A \\
4K & Flex & 0.49 & 16.25 & 0.40 & 343.79 \\
4K & Ours & 1.57 & \textbf{52.17} & 0.12 & 1,161.15 \\
\midrule
8K & SDPA & 1.09 & 36.40 & 5.73 & 95.91 \\
8K & SGLang & N/A & N/A & N/A & N/A \\
8K & Flex & 0.51 & 17.14 & 2.42 & 227.24 \\
8K & Ours & 1.66 & \textbf{55.36} & 0.50 & 1,098.13 \\
\midrule
16K & SDPA & 1.15 & 38.22 & 27.03 & 81.34 \\
16K & SGLang & N/A & N/A & N/A & N/A \\
16K & Flex & 0.50 & 16.67 & 22.95 & 95.80 \\
16K & Ours & 1.73 & \textbf{57.72} & 3.17 & 693.37 \\
\midrule
32K & SDPA & 1.15 & 38.48 & 110.43 & 79.65 \\
32K & SGLang & N/A & N/A & N/A & N/A \\
32K & Flex & 0.50 & 16.63 & 148.13 & 59.38 \\
32K & Ours & 1.76 & \textbf{58.79} & 15.35 & 573.05 \\
\midrule
64K & SDPA & 1.11 & 37.14 & 439.09 & 80.13 \\
64K & SGLang & N/A & N/A & N/A & N/A \\
64K & Flex & 0.50 & 16.56 & 768.34 & 45.79 \\
64K & Ours & 1.76 & \textbf{58.65} & 108.75 & 323.54 \\
\midrule
128K & SDPA & 1.12 & 37.17 & 1,774.27 & 79.32 \\
128K & SGLang & N/A & N/A & N/A & N/A \\
128K & Flex & 0.49 & 16.46 & 3,201.66 & 43.96 \\
128K & Ours & 1.77 & \textbf{58.87} & 517.15 & 272.14 \\
\bottomrule
\end{tabular}
\end{sc}
\end{small}
\end{center}
\vskip -0.1in
\end{table*}
\begin{table*}[b!]
\caption{Hardware and runtime configurations throughout the evaluation}
\label{tab:hardware}
\vskip 0.15in
\begin{center}
\begin{small}
\begin{sc}
\setlength{\tabcolsep}{1pt}
\begin{tabular}{lccr}
\toprule
 & \textbf{Intel Xeon 6767P} & \textbf{Intel Core Ultra 7 265K} \\
\midrule
Microarchitecture & Granite Rapids (server) & Arrow Lake-S (hybrid) \\
Sockets $\times$ cores & $2 \times 64$; one socket used & $1 \times 20$ (8 P-cores + 12 E-cores) \\
SMT & Enabled & Not supported \\
ISA & AVX-512, AMX-BF16 & AVX2, AVX-VNNI\\
Max. core frequency & 3.9 GHz & 5.4\,GHz (P), 4.6\,GHz (E) \\
L1D per core & 48\,KiB & 48\,KiB (P), 32\,KiB (E) \\
L2 & 2\,MiB  & 3\,MiB P-core; 4\,MiB shared E-cores \\
L3 & 336\,MiB & 30\,MiB \\
NUMA & SNC-2: 4 nodes, 2 per socket & 1 node \\
Main memory & 512GiB DDR5, 8 channels, 6400 MT/s & 48GiB DDR5, 2 channels, 8000 MT/s \\
\midrule
Threads used & 64, cores 0--63 (socket 0) or SNC Node 1 & 20, cores 0--19 (P and E) \\
Placement (kernel bench.) & \texttt{numactl} cores 0--63, nodes 0,1; & all cores, single node; \\
 & \texttt{OMP\_PLACES=cores}, \texttt{OMP\_PROC\_BIND=close} & same OpenMP binding \\
Placement (E2E bench.) & unpinned & unpinned \\
Allocator & tcmalloc (gperftools 2.18.1) & tcmalloc (gperftools 2.18.1) \\
OS / kernel & Ubuntu 24.04.1 & Ubuntu 26.04 LTS \\
Compiler / BLAS & GCC 13.3.0 / oneAPI MKL 2026.1 & GCC 15.2.0 / oneAPI MKL 2026.0 \\
PyTorch / SGLang / vLLM & 2.12.0 / v0.5.18 / v0.28.0 & 2.12.0 / v0.5.18 / v0.28.0 \\
\bottomrule
\end{tabular}
\end{sc}
\end{small}
\end{center}
\end{table*}
\begin{table*}[t!]
\caption{Kernel parameters adopted throughout the evaluation}
\label{tab:kernel-params}
\vskip 0.15in
\begin{center}
\begin{small}
\begin{sc}
\begin{tabular}{lccr}
\toprule
 & \textbf{Xeon 6767P} & \textbf{Core Ultra 7 265K} \\
\midrule
Q-block rows $q_r$ & 64 & 64 \\
KV-block columns $k_c$ (tile $T_{kv}$) & 512 & 512 \\
L3 capacity used by the model & 336 MiB & 30 MiB  \\
L3 budget fraction $f_{\mathrm{L3}}$ & 0.8 & 0.8 \\
L2 capacity used by the model & 2\,MiB (per-core L2) & 1\,MiB (E-core share of the 4\,MiB cluster L2) \\
L2 budget fraction $f_{\mathrm{L2}}$ & 0.65 & 0.65 \\
\bottomrule
\end{tabular}
\end{sc}
\end{small}
\end{center}
\end{table*}


\end{document}